\documentclass[conference]{IEEEtran}
\IEEEoverridecommandlockouts

\usepackage{cite}
\usepackage{amsmath,amssymb,amsfonts}
\usepackage{algorithmic}
\usepackage{graphicx}
\usepackage{textcomp}
\usepackage{xcolor}
\usepackage{multirow}
\usepackage{hhline}
\usepackage{booktabs}
\usepackage{array}
\usepackage{amsfonts}
\usepackage{amsmath}
\usepackage[hidelinks]{hyperref}

\newcommand{\etal}{\textit{et al}. } 

\def\BibTeX{{\rm B\kern-.05em{\sc i\kern-.025em b}\kern-.08em
    T\kern-.1667em\lower.7ex\hbox{E}\kern-.125emX}}
\begin{document}
\title{\textbf{ST-HCSS: Deep Spatio-Temporal Hypergraph Convolutional Neural Network for Soft Sensing}\\

\thanks{\textsuperscript{*}Corresponding Author: Junn Yong Loo (loo.junnyong@monash.edu)}

}

\DeclareRobustCommand*{\IEEEauthorrefmark}[1]{%
  \raisebox{0pt}[0pt][0pt]{\textsuperscript{\footnotesize #1}}%
}
\author{
    \IEEEauthorblockN{
        Hwa Hui Tew\IEEEauthorrefmark{1}, 
        Fan Ding\IEEEauthorrefmark{1},
        Gaoxuan Li\IEEEauthorrefmark{1},
        Junn Yong Loo\IEEEauthorrefmark{1,\textasteriskcentered},
        Chee-Ming Ting\IEEEauthorrefmark{1},
        Ze Yang Ding\IEEEauthorrefmark{2},
        Chee Pin Tan\IEEEauthorrefmark{2}
    }
    \IEEEauthorblockA{
        \IEEEauthorrefmark{1}School of Information Technology, Monash University Malaysia\\
        \IEEEauthorrefmark{2}School of Engineering, Monash University Malaysia\\ 
    }
}
\maketitle

\begin{abstract}
Higher-order sensor networks are more accurate in characterizing the nonlinear dynamics of sensory time-series data in modern industrial settings by allowing multi-node connections beyond simple pairwise graph edges. In light of this, we propose a deep spatio-temporal hypergraph convolutional neural network for soft sensing (ST-HCSS). In particular, our proposed framework is able to construct and leverage a higher-order graph (hypergraph) to model the complex multi-interactions between sensor nodes in the absence of prior structural knowledge. To capture rich spatio-temporal relationships underlying sensor data, our proposed ST-HCSS incorporates stacked gated temporal and hypergraph convolution layers to effectively aggregate and update hypergraph information across time and nodes. Our results validate the superiority of ST-HCSS compared to existing state-of-the-art soft sensors, and demonstrates that the learned hypergraph feature representations aligns well with the sensor data correlations. 
The code is available at \href{https://github.com/htew0001/ST-HCSS.git}{https://github.com/htew0001/ST-HCSS.git}

\end{abstract}


\section{Introduction}

As modern industrial processes advance and become increasingly complex, accurate measurement of dominant (key) variables plays a significant role in effective process monitoring, controlling, and optimization \cite{feng2020dual}. While dominant process variables such as temperature, pressure, flow rate, and density are of great importance in many industrial processes, acquiring them remains challenging due to the labor-intensive sensor deployment and Internet of Things (IoT) integration. Furthermore, physical IoT sensors are often susceptible to harsh environmental conditions, and measurement lags, in addition to requiring frequent maintenance. These limitations could leads to poor sensor measurements, significantly contributing to operational risks \cite{wei2020virtual}.
To circumvent the aforementioned issues, soft sensors have been widely adopted to estimate the hard-to-measure dominant variables, given the easy-to-measure auxiliary sensor variables \cite{jiang2020review}. Not only do soft sensors allows a more accurate characterization of the process, it also enhances fault detection and graceful degradation \cite{zeroshot}. In recent years, the proliferation of IoT systems enable the acquisition of operational data across various industries. As a result, the abundance of data information facilitate the development of data-driven soft sensors, and thereby eliminating the need for first principle model derived based on extensive domain expertise \cite{ddsr}.

\begin{figure*}
    \centering
    \includegraphics[width=1\linewidth]{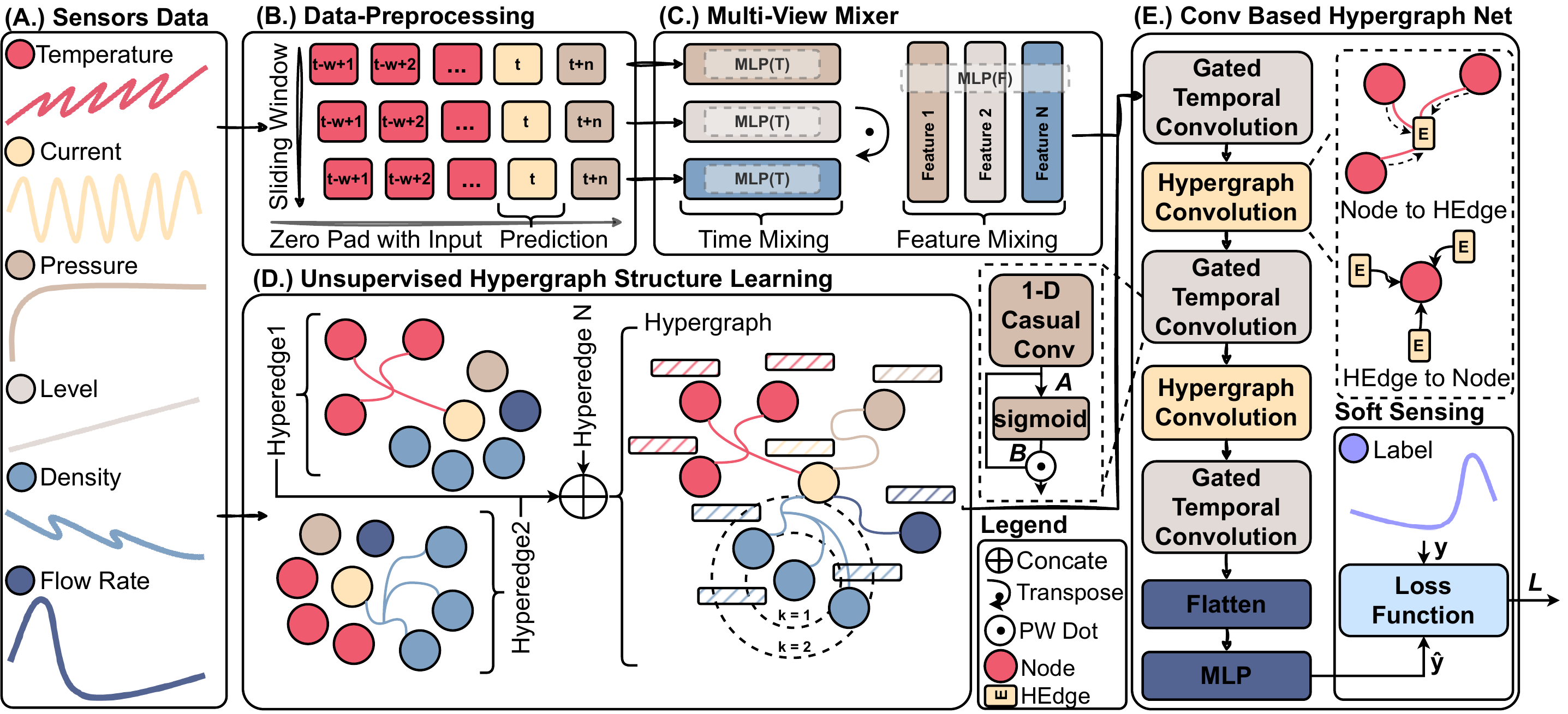}
    \vspace{-8mm}
    \caption{Overview of the designed architecture. It begins with the (A.) raw sensor data preprocessed by (B.) that transforms into a sliding window format. Next, a (C.) multi-view mixer that extracts spatio-temporal data features. Meanwhile, an (D.) unsupervised hypergraph structure learning module to classify related sensor nodes with hyperedges. Subsequently, a (E.) convolutional-based hypergraph representation learning module that produces the final prediction.}
    \label{fig:architecture}
    \vspace{-5mm}
\end{figure*}

Unlike statistical modeling methods, the advent of deep learning has revolutionized system modeling via learning deep hierarchical representation of data. 
In particular, convolutional neural networks (CNN) excel at capturing spatial features \cite{tim1,tim2}, while long short-term memory (LSTM) is effective in modeling temporal dependencies \cite{dlss-survey}. Recent state-of-the-art approaches such as weight-stacked autoencoder (VW-SAE) \cite{vwsae}, stacked target-related autoencoder (STAE) and gated STAE (GSTAE) \cite{gstae}, have further advanced the field by incorporating deep autoencoders to better retains model information across their multi-layer feature embeddings. 
Additionally, graph neural networks (GNN) have also been incorporated to model non-Euclidean relationships between sensor nodes. For examples, Huang \etal constructed a multi-modal GNN that leveraged sensors data and textual information for a better model representation \cite{gnn3}. Zhu \etal leveraged domain adaptation combined with GNN for soft sensor modeling \cite{zhu2023domain}. Feng \etal applied GNN for soft sensing to estimate the endpoint composition in steel \cite{gnn4}. Nevertheless, these GNN models neglect temporal aspects of the sensor data. This limits the model ability to fully leverage the rich spatio-temporal sensor dynamics, pivotal in high-fidelity soft sensing. 
To address this, Wang \etal combined 1D-CNN and GNN to construct a spatio-temporal network for better soft sensing performance \cite{wang2024soft}. Zhu \etal and Jia \etal extends this method to stacked 1D-CNN and GNN blocks \cite{zhu2022dynamic,gnn1}. 

Recently, hypergraph neural networks \cite{HGNN} have made immense progress in performing many industrial tasks, such as fault diagnosis \cite{yan2024multisensor}, predicting remaining useful life \cite{wu2024temporal}, and anomaly detection \cite{liang2021industrial}. However, pairwise edge connections in a graph are too restrictive to adequately represent the nonlinear interactions among system or sensor nodes. In contrast, higher-order connections inherent in hypergraph offer a more generalized representation of the complex sensor-to-sensor relationships.
To address the aforementioned issues, we propose a deep spatio-temporal hypergraph convolutional soft sensing (ST-HCSS) framework. 
To the best of our knowledge, this is the first soft sensing work based on spatio-temporal hypergraph. 
Our contributions are highlighted as follows: (1) We introduce a multi-view mixer to model the intersignal relationship (across time) and intrasignal relationship (across sensor nodes) in the data. (2) We incorporate stacked gated temporal convolution and hypergraph convolution layers to extract expressive latent spatial and temporal features. (3) Our results demonstrate that ST-HCSS achieves superior soft sensing performance on real industrial processes compared to state-of-the-art methods. Furthermore, we show that our learned hypergraph edge features are consistent with the data correlation. 

\section{METHODS}
\subsection{Problem Formulation}

Fig. \ref{fig:architecture} illustrates an overview of our proposed framework. Given a time-series sensor data with auxiliary variables $\mathbf{x}_{t} \in \mathbb{R}^{D}$ and dominant variable $\mathbf{y}_{t} \in \mathbb{R}^{Z}$ comprising $T$ observations. Our goal is to predict the dominant variable using the auxiliary variables by learning a deep soft sensing model. 
To effectively capture the intrinsic dynamic temporal dependencies of the data, a sliding window of size $W$ is applied to generate overlapping windows of time-series sensor input. Therefore, the soft sensing model is defined as $f: \mathbb{R}^{D \times W} \to \mathbb{R}^{Z}$, with soft sensor input (sliding-window) $\mathbf{X}_{t} = [\mathbf{x}_{t-W+1},\mathbf{x}_{t-W+2},\dots,\mathbf{x}_{t}] \in \mathbb{R}^{D \times W}$. Nevertheless, this model mapping does not account for the spatial dependencies between sensor nodes. To incorporate non-Euclidean spatial relationships, we represent the input data as a hypergraph $\mathcal{H}$ = $(\mathcal{V},\mathcal{E})$, where the set of nodes $ v_i \in \mathcal{V}$ is the sensors with node features $\mathbf{X}_{t}^{i}$, where $\mathbf{X}^{i}$ denotes the $i^{\text{th}}$ row of matrix $\mathbf{X}$. The set of hyperedges $\mathcal{E}$ is to be determined via unsupervised structure learning.

\subsection{Unsupervised Hypergraph Structure Learning}
In practice, a topological structure that characterizes the interplay between the monitored process (sensor) nodes of a industrial process is often unknown or requires expert domain knowledge to determine. In light of this, we learn this topological structure underlying the industrial system by constructing a weighted hypergraph $\mathcal{G}$ = $(\mathcal{V},\mathcal{E},\mathbf{W})$ where the relationships between sensor nodes $ v\in\mathcal{V}$ are governed by the set of hyperedges $e\in\mathcal{E}$, and $\mathbf{W}$ denotes the weight of the connections between node $i$ and $j$. In contrast to the simple graph, which represents the topological structure using an adjacency matrix, $\mathbf{A} \in \mathbb{R}^{|\mathcal{V}| \times |\mathcal{V}|}$, the hypergraph is characterized by the incidence matrix $\mathbf{H} \in \mathbb{R}^{|\mathcal{V}| \times |\mathcal{E}|}$. Here, we calculate the Euclidean distance between nodes as $\mathcal{D}(v_{i}, v_{j}) = \|v_{i} - v_{j}\|$ and the average pairwise distance as $\triangle = \frac{1}{|\mathcal{V}|} \sum_{i} \mathcal{D}(v_{i}, v_{j})$. To construct the hyperedges using these distances, we perform $k$-nearest neighbour (KNN) on each node $v_{j}$ to obtain the $k$-nearest nodes $v_{i}$ i.e., $v_{i} \in \text{KNN}(v_{j})$. The hyperedges $h(v_i,e_j)$ and the hyperedge weights $w(v_i,e_j)$ are then identified as follows:
\begin{align}
h(v_i,e_j) &= 
\begin{cases}
1, & \text{if } v_{i} \in \text{KNN}(v_{j}) \\
0, & \text{otherwise}
\end{cases}
\label{eq:1}
\\
w(v_i,e_j) &= 
\begin{cases}
\text{exp}\left(-\frac{\mathcal{D}(v_{i}, v_{j})^{2}}{\triangle}\right), & \text{if } v_{i} \in \text{KNN}(v_{j}) \\
0, & \text{otherwise}
\end{cases}
\label{eq:2}
\end{align}
where $h(v_i,e_j)$ is the element $\mathbf{H}_{ij}$ of the incidence matrix $\mathbf{H}$, and $w(v_i,e_j)$ is the element $\mathbf{W}_{ij}$ of the hyperedges weight matrix $\mathbf{W}$.
The degree of vertex $d(v) \in \mathbb{R}$ is defined as the summation of all hyperedges weight attached to vertex $v$, where $d(v) = \sum_{e \in \mathcal{E}} h(v,e)$ that is stored in a diagonal matrix $\mathbf{D}_{v} = \mathbb{R}^{|v|\times|v|}$. Similarly, the degree of a hyperedge $d(e)$ is defined as the sum of the incidence of all vertex across hyperedges $e$, where $d(e) = \sum_{v \in \mathcal{V}} h(v,e)$ that is stored in a diagonal matrix $\mathbf{D}_{e} = \mathbb{R}^{|e|\times|e|}$.

\subsection{Convolution Based Hypergraph Representation Learning}

To effectively capture the spatio-temporal soft sensing characteristics, we first introduce a multi-view mixer to perform global feature extraction (mixing) across the spatial and time dimensions (views) \cite{mlp}. Subsequently, we incorporate dynamic hypergraph convolutional, in which we stack the temporal and spatial (hypergraph) convolution blocks to effectively encode long-term spatio-temporal relationships. The proposed model consists of three main components: multi-view mixer, gated temporal convolution, and  hypergraph convolution, as illustrated in Fig. \ref{fig:architecture}. \\ \textbf{Multi-View Mixer:} Inspired by the work \cite{mlp}, we propose two MLP-mixing models that  take into account the different views in our time-series soft sensor: time-mixing MLP and feature-mixing MLP to model the transformation between multi-view information effectively. Time mixing is performed across the timesteps of the sliding window input $\mathbf{X}$, where we apply a single-layer
perceptron $(\textbf{SLP}_{T})$ that is applied in parallel to all node features; feature mixing is performed on the column by $\mathbf{X}^{T}$, where we apply two multi-layer perceptron $(\textbf{MLP}_{P})$ that is applied in parallel to all timesteps. Finally, the outputs from the time and feature mixing MLPs are combined via a skip connection \cite{skip}. The overall multi-view mixer module can be expressed as follows:
\begin{align}
\begin{split}
\noindent \mathbf{U}_{i,*}=\mathbf{X}_{i,*}+\sigma_r\big[\textbf{W}_{1}(\mathrm{LayerNorm}(\mathbf{X})_{i,*})+\textbf{b}_{1}\big]
\end{split} \\
\begin{split}
\noindent \mathbf{Y}_{*,j}=\mathbf{U}_{*,j}+\big[\textbf{W}_{3}\big(\sigma_r\big[\textbf{W}_{2}(\mathrm{LayerNorm}(\mathbf{U})_{*,j})+\textbf{b}_{2}\big]\big)+\textbf{b}_{3}\big]
\end{split}
\end{align} 
where $i=1\hdots \text{R}$ and $j=1\hdots \text{C}$; R denotes the rows and C the columns of $\mathbf{X}$. Here, $\textbf{W}_{1},\textbf{W}_{2},\textbf{W}_{3}$, $\textbf{b}_{1},\textbf{b}_{2},\textbf{b}_{3}$ are learnable weights and biases, and $\sigma_r$ is the ReLU activation function. This multi-view mixer module extracts global features across both time and node features in a complex multivariate dataset.
\\  
\textbf{Gated Temporal Convolution:} 
We apply the gated temporal convolution (GTC) on sensors data that often exhibit strong time coherence as depicted in Fig.\ref{fig:architecture}. Specifically, we consider a causal temporal convolution in which local temporal convolution attends exclusively to features from the preceding timesteps. Additionally, we incorporate a gated mechanism to improving the model's ability in capturing complex temporal correlation via selective information flow, and to aid in retention of long-term dependencies.
The one-dimensional temporal convolution is defined as follows:
\begin{align}
\begin{split}
(\mathbf{w} \star_{d} \mathbf{x})_{t} = \sum_{\tau=0}^{K - 1} \mathbf{w}_{\tau} \cdot \mathbf{x}_{t-\tau d} + \textbf{b}_{\tau}
\end{split}
\end{align}
where $\mathbf{w} \in \mathbb{R}^{K}$ is the convolution kernel of size $K$, and $\mathbf{x} \in \mathbb{R}^{W}$ is taken as the row of the input feature $\mathbf{X}_{t} = [\mathbf{x}_{t-W+1},\mathbf{x}_{t-W+2},\dots,\mathbf{x}_{t}]$. While a larger kernel size $K$ results in a larger receptive field better at capturing long-range dependencies, an excessively large $K$ can overlook the importance of local features.
A gated temporal convolution block then consists of a stack of convolution layers, with each layer $k$ performed as follows:
\begin{align}
\begin{split}
\mathbf{h}^{k+1} = (\mathbf{w}_{f}^{k} \star \mathbf{x}^{k})_{t} \odot \sigma_s(\mathbf{w}^{k}_{g} \star \mathbf{x}^{k})_{t}
\end{split}
\end{align}
where the convolution with respect to the gate kernel $\mathbf{w}_{g}(t)$ is passed through sigmoid activation $\sigma_s$. The output is then combined with another convolution with respect to the filter kernel $\mathbf{w}_{f}(t)$ via element-wise product $\odot$.
\\ 
\textbf{Spectral Hypergraph Convolution:}
Hyperedges provide an accurate characterization of the complex spatial correlations between system or sensor nodes in industrial processes. In light of this, soft sensing can then be formulated as a hypergraph node regression problem. 
In particular, a spectral convolution of the hypergraph features $\mathbf{x} \in \mathbb{R}^{N}$ with respect to filter $\mathbf{g}: \mathbb{R} \to \mathbb{R}$ can be formulated as follows:
\begin{align}
\begin{split}
\mathbf{g} \star \mathbf{x} 
=\boldsymbol{\Phi}g(\boldsymbol{\Lambda})\boldsymbol{\Phi}^Tx  
\label{eq:4}
\end{split}
\end{align}
where $g(\boldsymbol{\Lambda}) = \text{diag}(\mathbf{g}(\lambda_{1}),\mathbf{g}(\lambda_{2}),\dots,\mathbf{g}(\lambda_{n}))$, and $\boldsymbol{\Phi}$ and $\boldsymbol{\Lambda}$ are obtained via the eigen-decomposition of the normalized hypergraph Laplacian matrix \cite{HProb}:
\begin{align}
\begin{split}
\mathcal{L} = \mathbf{I} - \mathbf{N} = \boldsymbol{\Phi}\boldsymbol{\Lambda}\boldsymbol{\Phi}^T
\end{split}
\end{align}
where 
$\mathbf{N} = \mathbf{D}_v^{-1/2}\mathbf{H}\mathbf{W}\mathbf{D}_e^{-1}\mathbf{H}^T\mathbf{D}_v^{-1/2}$ is the normalized weighted hypergraph adjacency.
Given a hypergraph features $\mathbf{X}^{l} \in \mathbb{R}^{N \times C^{l}}$, 
the spectral convolution operation for each $l^{\text{th}}$ layer can then be formulated \cite{hypergraph} as
\begin{align}
\begin{split}
\mathbf{X}^{l+1} &=\sigma_r\left(\mathbf{D}_v^{-1/2}\mathbf{H}\mathbf{W}\mathbf{D}_e^{-1}\mathbf{H}^T\mathbf{D}_v^{-1/2}\mathbf{X}^l\mathbf{\Theta}^l\right)
\label{eq:6}
\end{split}
\end{align}
where $\sigma_r$ denotes the ReLU activation and $\Theta^{l} \in \mathbb{R}^{C^{l} \times C^{l+1}}$ is the set of learnable filter parameters. 
\begin{table*}[tb]
\centering
\caption{Quantitative results of ST-HCSS across three different dominant process variables (and their position).Lower NRMSE, NMAE, MAPE indicate better results, higher $\text{R}^2$ shows better model fit to the MFP data.}
\vspace{-3mm}
\label{tab:regression_metrics}
\resizebox{\textwidth}{!}{%
\begin{tabular}{@{}r|cccc|cccc|cccc@{}}
\toprule
\multicolumn{1}{c|}{\multirow{2.5}{*}{Methods}} &
  \multicolumn{4}{c|}{Pressure in 3-phase separator (PT501)} &
  \multicolumn{4}{c|}{Flow Rate Input Air (FT305)} &
  \multicolumn{4}{c}{Position of Valve (VC501)} \\ 
  \cmidrule(lr){2-5} \cmidrule(lr){6-9} \cmidrule(l){10-13} 
  & NMAE 
  & NRMSE 
  & MAPE 
  & $\mathrm{R}^2$ 
  & NMAE 
  & NRMSE 
  & MAPE 
  & $\mathrm{R}^2$ 
  & NMAE 
  & NRMSE 
  & MAPE 
  & $\mathrm{R}^2$  \\ 
\midrule
SVR \cite{svr}  
& 2.418 & 4.290 & 0.409 & 0.761
& 4.137 & 5.277 & 3.360 & 0.962
& 4.604 & 6.758 & 2.239 & 0.833 \\
PLSR \cite{plsr}
& 4.615 & 5.682 & 0.780 & 0.581  
& 4.916 & 6.002 & 3.955 & 0.957 
& 5.077 & 7.676 & 2.455 & 0.783\\
DNN \cite{DNN}  
& 3.033 & 4.539 & 0.515 & 0.731 
& 3.573 & 4.872 & 2.839 & 0.970 
& 4.925 & 6.842 & 2.433 & 0.828 \\
VW-SAE \cite{vwsae} 
& 2.978 & 4.415 & 0.502 & 0.746
& 3.168 & 4.481 & 2.544 & 0.973 
& 4.405 & 5.893 & 2.315 & 0.872 \\
STAE \cite{gstae} 
& 2.737 & 4.420 & 0.464 & 0.746
& 3.046 & 4.030 & 2.492 & 0.978 
& 4.413 & 6.128 & 2.147 & 0.862 \\
GSTAE \cite{gstae} 
& 2.404 & 3.926 & 0.407 & 0.800 
& 2.865 & 3.833 & 2.368 & 0.980 
& 3.998 & 5.770 & 1.949 & 0.878 \\
ST-GNN \cite{gnn} 
& 2.399 & 3.867 & 0.402 & 0.805
& 2.854 & 3.712 & 2.351 & 0.981 
& 3.652 & 4.717 & 1.647 & 0.905  \\
{HGNN} \cite{HGNN} 
& {2.428} & {3.889} & {0.298} & {0.898} 
& {3.326} & {4.637} & {2.548} & {0.972} 
& {2.608} & {3.664} & {1.250} & {0.952} \\
\midrule
\textbf{ST-HCSS (Ours)} 
& \textbf{1.975} & \textbf{3.647} & \textbf{0.242} & \textbf{0.910} 
& \textbf{2.441} & \textbf{3.492} & \textbf{1.908} & \textbf{0.984} 
& \textbf{2.298} & \textbf{3.544} & \textbf{1.096} & \textbf{0.955} \\
\bottomrule
\end{tabular}%
}
\vspace{-5mm}
\end{table*}

Finally, we flatten and apply a MLP readout layer on the output features of the convolution blocks to obtain our final soft sensor prediction $\hat{y}^{(t)}$. 
For model training, we employ the mean squared error (MSE) loss function:
\begin{equation}
\begin{aligned}
L_{\text{MSE}}=\frac{1}{T}\sum_{t=0}^{T}\left(y^{(t)}-\hat{y}^{(t)}\right)^{2}
\end{aligned}
\end{equation}
with ground truth $y$ and train the ST-HCSS network models in an end-to-end fashion.

\section{EXPERIMENTAL RESULTS}

\label{sec:pagestyle}
\subsection{Real-World Case Study:}
To validate the performance of the proposed ST-HCSS model, we conduct experiments on a real-world industrial process, i.e., the Cranfield Multiphase Flow Process (MFP) facility \cite{mfp}. This MFP facility consists of 24 distinct process (sensor) variables, sampled at a frequency of 1 Hz on normal operations 
The process is operated and varied across 20 different setpoints to capture a wide range of operational scenarios and process dynamics.
\subsection{Experimental Setup} 
\noindent \textbf{Baselines:} Our proposed ST-HCSS model is compared against eight different soft sensor model baselines.
These baselines include support vector regression (SVR) \cite{svr}, partial least square regression (PLSR) \cite{plsr}, Dense Neural Network (DNN) \cite{DNN}, variable-weighted stacked autoencoder (VW-SAE) \cite{vwsae}, stacked target-related autoencoder (STAE) and its gated variant (GSTAE) \cite{gstae}, graph neural network (ST-GNN) \cite{gnn} and hypergraph neural network (HGNN) \cite{HGNN}. 
\\ \textbf{Evaluation Metrics:} 
We assess the model performance using the evaluation metrics: NMAE, NRMSE, MAPE, and $\text{R}^2$, commonly used for validating regression models. 
\begin{figure}
    \centering
    \includegraphics[width=1\linewidth]{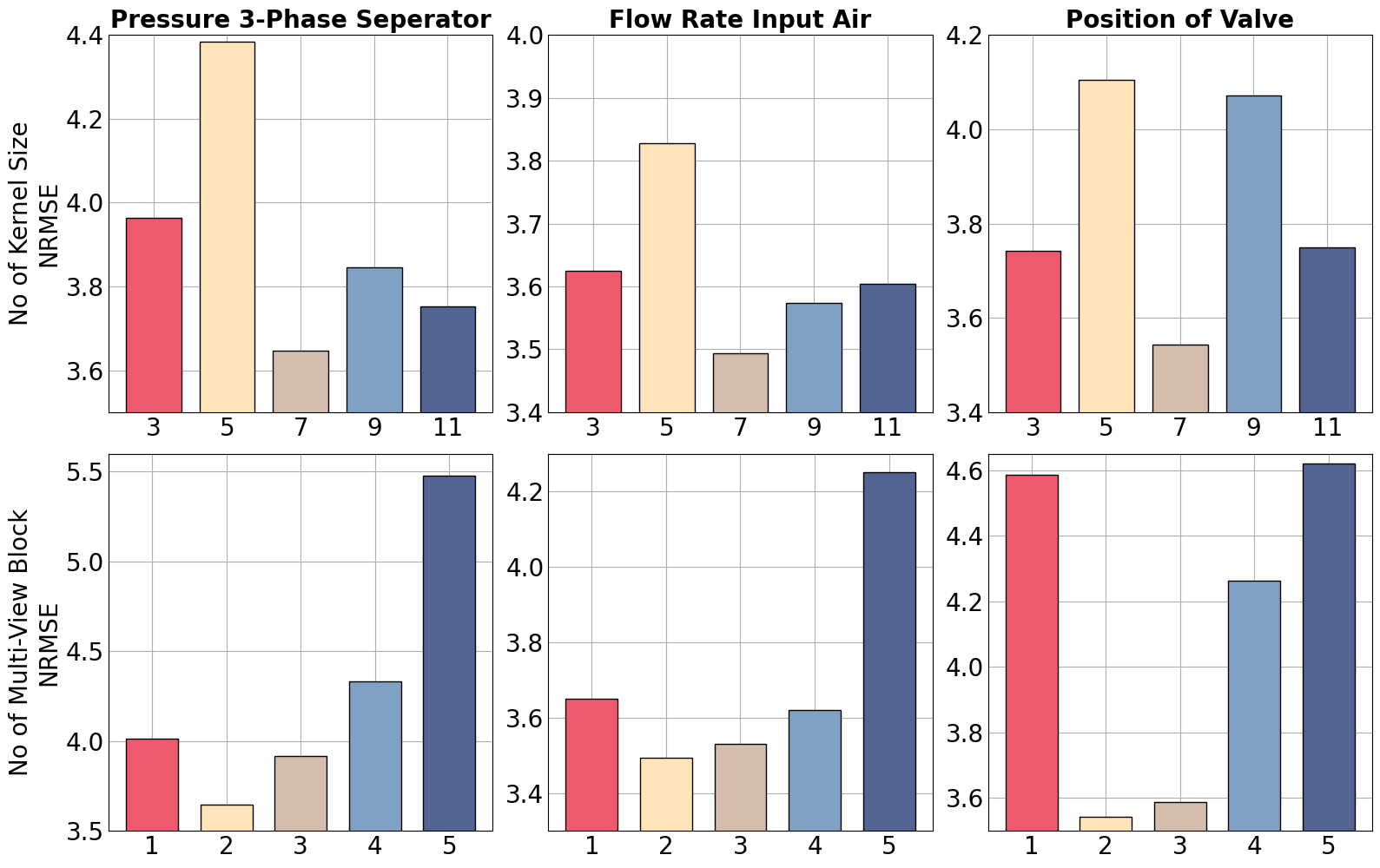}
    \caption{Hyperparameter analysis on ST-HCSS. First row shows kernel size of \{3,5,7,9,11\}. Second row represents mixer block number of \{1,2,3,4,5\}.}
    \label{fig:hyperparameteranalysis}
    \vspace{-3mm}
\end{figure}

\begin{figure}
    \centering
    \includegraphics[width=1\linewidth]{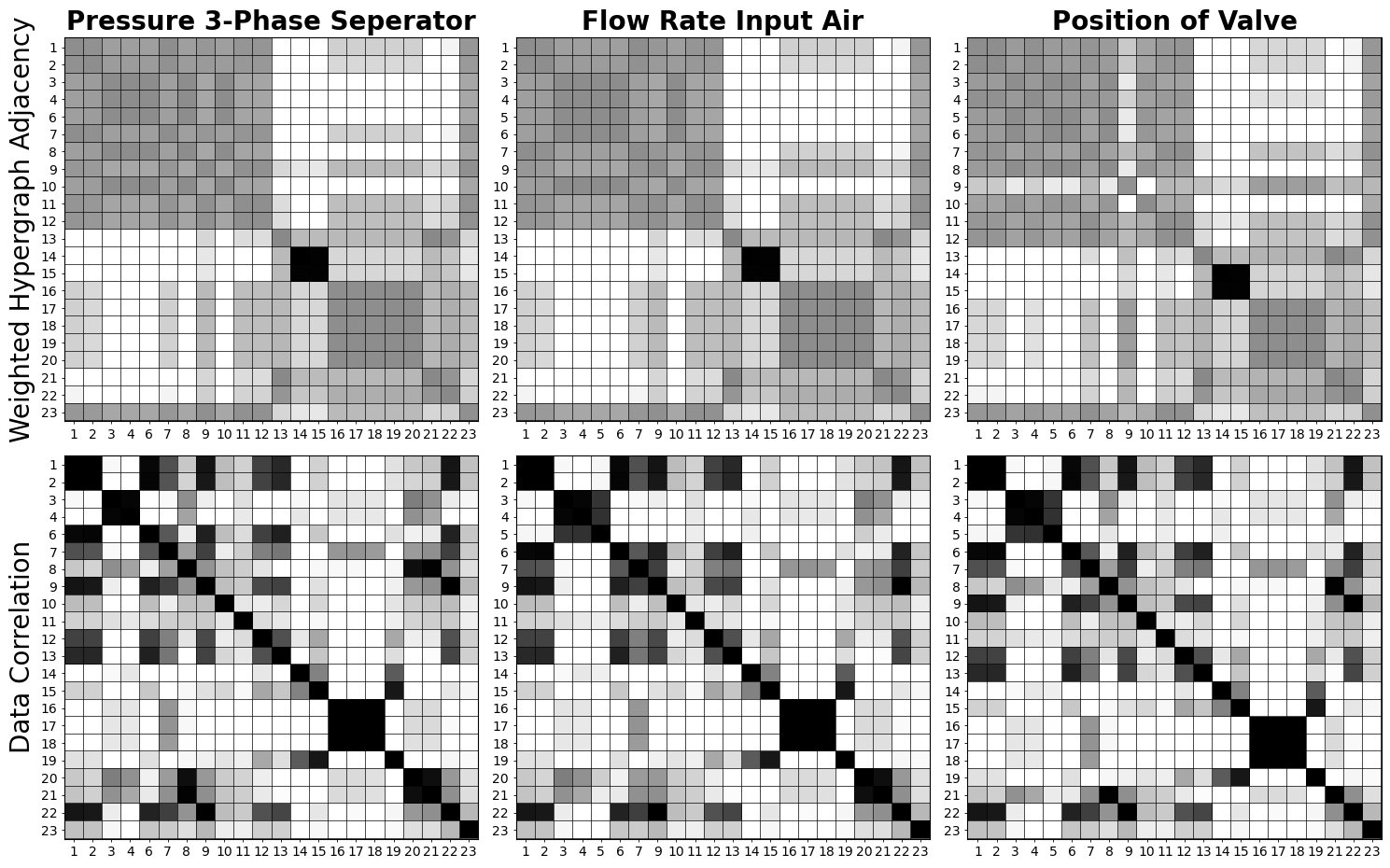}
    \caption{Heatmap of normalized weighted hypergraph adjacency and data correlation with respect to each of the process in table I }
    \label{fig:Visualization}
    \vspace{-3mm}
\end{figure}

\noindent \textbf{Implementation Details:} The data are partitioned into training, validation, and testing sets with a split ratio of 60-20-20. Throughout the model training, we set a batch size of 64, a sliding window size of 85, and a dropout of 0.2 to prevent overfitting. Also, the number of multi-view mixers is set to 2, and the kernel size and dilation of GTC are set to 7 and 1, respectively. The model is trained for 200 epochs on Adam optimizer \cite{adam} at a learning rate of 0.001. 
\\ \textbf{Overall Performance:} Table \ref{tab:regression_metrics} compares the performance of ST-HCSS to the baselines on three different process variables. In overall, the results show that ST-HCSS consistently outperformed all the baselines across every metrics in predicting the three dominant process variables. In particular, our model significantly outperforms the ST-GNN and HGNN, neither of which incorporate the proposed multi-view mixer and  gated temporal and hypergraph convolutions. This exemplifies the capability of ST-HCSS and the importance of its mixer and convolution modules in effectively extracting salient spatio-temporal sensor characteristics for high-fidelity soft sensing.
\\
\textbf{Hyperparameter Analysis:} To further investigate the influence of model hyperparameters on the multi-view mixer and gated temporal convolution. Our results in Fig. \ref{fig:hyperparameteranalysis} show that the kernel size of 7 gives the best performance. In particular, a lower or higher kernel size introduces noise to the local convolutions and impacts soft sensing accuracy. Also, the ablation study shows that 2 multi-view mixer blocks give the best results. Having more mixer blocks leads to overfitting and compromises model generalization.  
\\ \textbf{Structural Analysis:} Fig. \ref{fig:Visualization} compares the normalized weighted hypergraph adjacency to the ground-truth data correlation for the three dominant variables. Consistent with the data correlation, the weighted hypergraph adjacency exhibit dense edge connections within sensor groups: pressures (1-7), flow rates (8-12), valve positions (20-22), in contrast to the sparse connections across different sensor groups. 

\section{Conclusion}
In this paper, we developed a ST-HCSS model that outperforms existing soft sensing models, demonstrating its ability in extracting complex higher-order spatial correlations between process variables. 

\section*{Acknowledgement}
This work was supported by the Advanced Computing Platform (ACP), Monash University Malaysia.
The work of Junn Yong Loo and Ze Yang Ding is supported by Monash University under the SIT Collaborative Research Seed Grants 2024 I-M010-SED-000242 and the SOE Collaborative Research Seed Grants 2024 I-M010-SED-000248, respectively.


\vspace{12pt}

\bibliographystyle{References/IEEEtran1}
\bibliography{References/mybibfile}

\end{document}